\documentclass[letterpaper, 10 pt, conference]{ieeeconf}  
\usepackage[utf8]{inputenc}
\usepackage{url}

\IEEEoverridecommandlockouts                              

\usepackage{cite}
\usepackage{graphicx}   %
\usepackage{caption}    %
\usepackage{subcaption} %
\usepackage{amsmath}

\title{\LARGE \bf
VacuumVLA: Boosting VLA Capabilities via a Unified Suction and Gripping Tool for Complex Robotic Manipulation
}

\author{
  Hui Zhou\textsuperscript{1,*,\dag},
  Siyuan Huang\textsuperscript{2,*},
  Minxing Li\textsuperscript{3,*},
  Hao Zhang\textsuperscript{1},
  Lue Fan\textsuperscript{3},
  Shaoshuai Shi\textsuperscript{4}
  \thanks{*Equal contribution.}
  \thanks{\dag Project leader.}
  \thanks{\textsuperscript{1}The Chinese University of Hong Kong, Hong Kong SAR, China}
  \thanks{\textsuperscript{2}Shanghai Jiao Tong University, Shanghai, China}
  \thanks{\textsuperscript{3}Institute of Automation, Chinese Academy of Sciences, Beijing, China}
  \thanks{\textsuperscript{4}DiDi Global, China}
}

\begin{document}

\maketitle
\thispagestyle{empty}
\pagestyle{empty}

\begin{abstract}

Vision-Language-Action (VLA) models have significantly advanced general-purpose robotic manipulation by harnessing large-scale pre-trained vision and language representations. Among existing approaches, a majority of current VLA systems employ parallel two-finger grippers as their default end-effectors. However, such grippers face inherent limitations in handling certain real-world tasks—such as wiping glass surfaces or opening drawers without handles—due to insufficient contact area or lack of adhesion.  

To overcome these challenges, we present a low-cost, integrated hardware design that combines a mechanical two-finger gripper with a vacuum suction unit, enabling dual-mode manipulation within a single end-effector. Our system supports flexible switching or synergistic use of both modalities, expanding the range of feasible tasks. We validate the efficiency and practicality of our design within two state-of-the-art VLA frameworks: DexVLA and $\pi_0$. Experimental results demonstrate that with the proposed hybrid end-effector, robots can successfully perform multiple complex tasks that are infeasible for conventional two-finger grippers alone. All hardware designs and controlling systems will be released. 


\end{abstract}

\section{INTRODUCTION}

Thanks to advances in vision-language models\cite{beyer2024paligemma, bai2025qwen2} and the accumulation of large-scale manipulation datasets\cite{o2024open, khazatsky2024droid}, embodied AI has made significant progress in recent years. By training on vast amounts of aligned visual and textual data, VLMs have acquired strong generalization capabilities and transferable representations, enabling robots to rapidly adapt to new tasks with little or even no task-specific supervision. At the same time, large-scale manipulation datasets provide rich examples of action demonstrations. Through imitation learning, robots can efficiently learn from these examples and master complex motor skills.

Building on recent progress, OpenVLA \cite{kim2024openvla} unifies a pretrained vision encoder, a pretrained large language model (LLM), and an action prediction module into a cohesive Vision-Language-Action (VLA) framework, enabling more intelligent and adaptable robotic manipulation. By leveraging the rich, general-purpose representations learned by the pretrained models, OpenVLA can generalize to novel tasks and achieve zero-shot task transfer through natural language instructions alone, without requiring task-specific fine-tuning. Further advancements are exemplified by $\pi_0$\cite{black2024pi_0} and $\pi_{0.5}$\cite{intelligence2025pi_5}, which incorporate more powerful VLM backbones and a flow-based action decoder, achieving robust and high-frequency control across a variety of robotic tasks.

However, effective complex manipulation requires more than semantic understanding and spatial reasoning—it is ultimately constrained by physical hardware, particularly the design and capabilities of the end effector. Current Vision-Language-Action (VLA) models predominantly rely on visual and linguistic inputs for task interpretation and planning, yet often overlook the end effector as a critical modality for real-world interaction. This sensory-motor asymmetry limits a robot’s ability to perform diverse and dexterous physical tasks, underscoring the need for co-design of intelligent control strategies and versatile hardware.

\begin{figure}[t]
    \centering
    \includegraphics[width=0.47\textwidth]{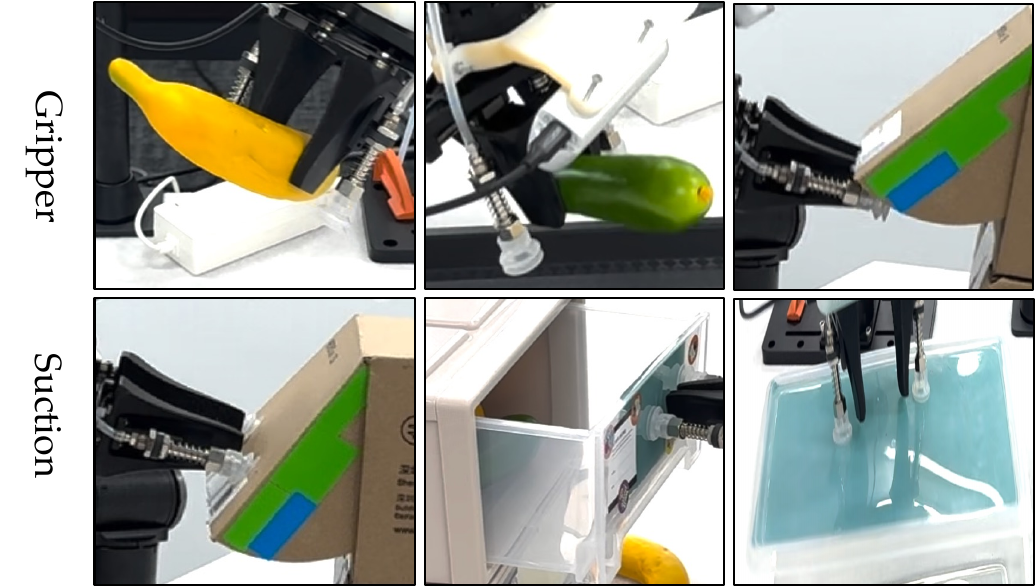}
    \caption{Illustration of the our end-effector with integrated gripper and suction cup.}
    \label{fig:fig1}
\end{figure}

To address these limitations, we develop a low-cost end effector that integrates suction and gripping functionalities, as shown in Fig. \ref{fig:fig1}, and complement it with a dedicated control and data collection system. To evaluate the effectiveness of our design, we conduct experiments using two distinct Vision-Language-Action (VLA) frameworks. The results demonstrate that our integrated end effector enables robots to successfully perform a range of household tasks that are previously unachievable with conventional grippers. Our key contributions are as follows:

\begin{itemize}
\item We develop a novel, low-cost end effector that combines suction and gripping capabilities. This hybrid design enables robots to perform challenging household tasks—such as opening handleless drawers—while maintaining strong performance in standard grasping operations.
\item We establish and validate a comprehensive data acquisition and control system. To demonstrate its effectiveness, we design and execute four distinct tasks: clearing a tabletop, opening a handleless plastic container, opening a handleless drawer, and opening a delivery cardboard box. We successfully validate our system using two open-source Vision-Language-Action (VLA) frameworks—DexVLA \cite{wen2025dexvla} and $\pi_0$ \cite{black2024pi_0}—achieving consistent and promising results in both.
\end{itemize}

\section{Related Works}

\subsection{End Effector Design}
\label{sec:related_effectors}

In grasping tasks, end-effector design is a crucial aspect that determines the upper limit of the system's capability to handle objects. Among these, parallel grippers are a very common type of end-effector, which grasp objects using driven jaws that can open and close. This type of gripper can handle various object shapes and properties, and its grasping capability can be enhanced by integrating additional sensors, such as tactile \cite{yuan2017gelsight, liu2022gelsight} and force \cite{li2025construction} sensors. Multi-finger grippers and dexterous hands are also popular end-effectors. Previous studies have explored different designs for multi-finger grippers \cite{shao2020unigrasp, burgess2025grasp, cutler2024benchmarking} and dexterous hands \cite{weng2025bidexhand, shaw2023leap, romero2024eyesight}, with some work specifically dedicated to developing tailored algorithms \cite{xu2023unidexgrasp, wan2023unidexgrasp++, lan2023dexcatch} and datasets \cite{liu2024realdex}. 

In contrast, vacuum suction grippers \cite{eppner2016lessons, schwarz2017data} grasp objects by generating suction force through vacuum pressure, enabling them to accomplish tasks that conventional grippers cannot perform. However, their effectiveness is restricted to objects with flat and smooth surfaces that can be sealed by the suction cup, such as glass. Moreover, vacuum suction grippers struggle to grasp porous or cloth-like objects effectively, as they cannot generate sufficient suction force in such cases. 

Beside single-function grippers, 
Multi-functional grippers \cite{d2023multimodal,um2023rec,zeng2018robotic,zeng2022robotic,son2025corner} which have proven effective in various challenges have been introduced to overcome the limitations of single-function grippers. The MIT-Princeton team \cite{zeng2022robotic} took 1st place in the stowing task at the 2017 Amazon Robotics Challenge with their integrated suction-grasping hardware. Recently, the champion \cite{son2025corner} of the 9th Robotic Grasping and Manipulation Competition (RGMC) held at ICRA 2024 also utilized an integrated suction-grasping hardware design.

\subsection{Vision Language Action Models}

Recent research on Vision-Language-Action (VLA) models \cite{kim2024openvla, black2024pi_0, intelligence2025pi_5, wen2025dexvla, yu2025forcevla, liurdt, shukor2025smolvla, brohan2022rt, zitkovich2023rt, cheang2024gr} has primarily focused on leveraging large-scale multimodal pretraining to achieve policy generalization across multiple instructions and diverse embodiments. These models can typically be decomposed into two modules: the first is a multimodal model responsible for encoding visual and linguistic information into tokens, and the second is an action expert that maps the encoded features directly to low-level control signals. The first module often incorporates reasoning mechanisms \cite{intelligence2025pi_5, wen2025dexvla} to improve instruction understanding. Currently, two common architectures are used for the action expert: flow-matching \cite{lipman2022flow} based and diffusion \cite{ho2020denoising} based. Flow-matching based architectures, such as $\pi_0$ \cite{black2024pi_0}, combine pretrained vision-language encoders with fast action decoders to enable high-frequency action outputs. Diffusion based models \cite{wen2025dexvla} support diverse and long-horizon behaviors through stochastic generation, but usually entail high training and inference costs. Despite significant progress, most existing VLA methods remain limited to visual and language inputs and are heavily constrained by hardware limitations.

\section{Methods}

In this section, we first introduce a multifunctional gripper that combines suction and finger grasping capabilities, enabling it to handle a wide variety of objects in household tasks. Second, we present our vision-language-action model, which is primarily enhanced by incorporating a new dimension for the suction tool.

\begin{figure}[t]
    \centering
    \includegraphics[width=0.5\textwidth]{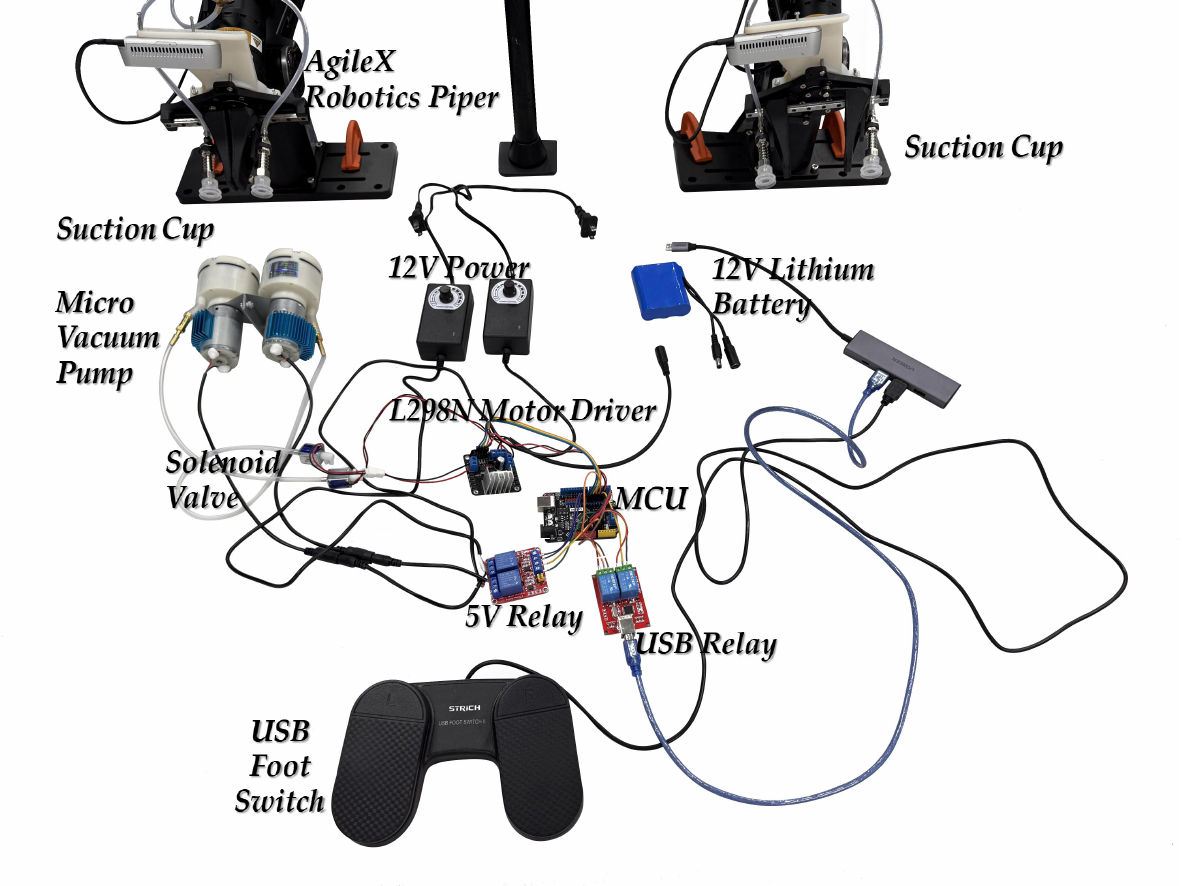}
    \caption{Hardware Details.}
    \label{fig:hardware}
\end{figure}

\begin{figure*}[htbp]
  \centering
  \includegraphics[width=\textwidth]{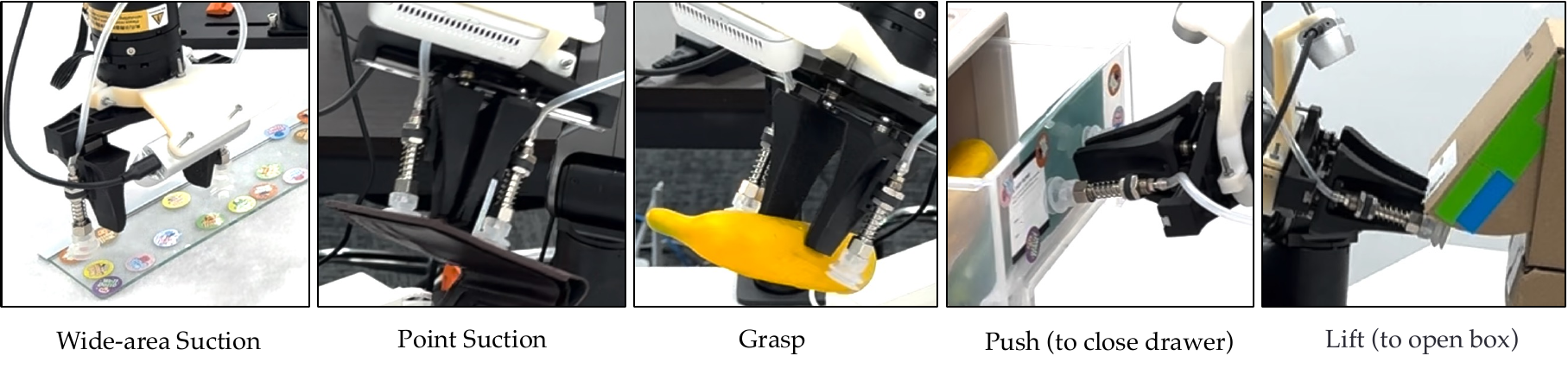}
  \caption{Prime actions: Adjustable gripping width for varying object sizes (first two). The remaining three primitives are based on the standard two-finger gripper.}
  \label{fig:prime_actions}
\end{figure*}

\subsection{Limitations of Previous VLA End-effectors}
\label{sec:limitations_of_previous}

As discussed in Section~\ref{sec:related_effectors}, the design of end-effectors is one of the key issues in the field of robotics. The parallel gripper is the most popular choice for VLAs. 

Parallel grippers are effective and easy to control, but due to their simple structure, they are unable to perform some relatively complex tasks. They are limited by object size, as they require sufficient space to form a stable grasp with their jaws. Furthermore, the grasping stability for certain objects cannot be guaranteed during manipulation, for example, a table tennis ball. Dexterous hands, on the other hand, are a comprehensive imitation of the human hand. Although capable of handling more complex tasks, they are more difficult to control due to the high degree-of-freedom (DoF). Furthermore, they still struggle with many household tasks, including opening a lid or drawer without a handle. 

Table~\ref{table:challenging_tasks} lists some typical tasks that parallel grippers and dexterous hands struggle to perform.

\begin{table}[htbp]
    \centering
    \caption{Typical examples of challenging tasks that popular end-effectors struggle to perform. These tasks can be accomplished by our hybrid end-effector.}
    \begin{tabular}{cl}
        \hline
        ID & Description \\
        \hline
        1 & Picking up an extremely thin piece of paper or glass \\
        2 & Holding a large object whose size exceeds the effector's max range  \\
        3 & Opening a lid or drawer without a handle \\
        4 & Open a closed cardboard box \\
        \hline
    \end{tabular}
    \label{table:challenging_tasks}
\end{table}

\subsection{Gripper Designs}
Based on the above analysis, we propose an innovative hardware design to overcome the limitations. The goal of the first step for our end effector is to meet the requirements of common household tasks. To this end, we target a variety of different household objects: elongated glass items, banana props, cucumber props, wallets, sealed plastic containers, handleless drawers, and delivery cardboard boxes. We define prime actions that are complementary to each other in terms of utility across different object types and scenarios, and ensured successful execution through teleoperation, guaranteeing that at least one primitive can successfully complete the task.


\subsubsection{Hardware Details}

The overall hardware visualization is shown in Fig. \ref{fig:hardware}.
Each specific component and the system control is described as follows.

\begin{itemize}
    \item \textbf{Two-Finger Gripper Base:} AgileX Robotics Piper~\cite{agilex_piper}
    \item \textbf{Micro Vacuum Pump:} 
    \begin{itemize}
        \item Voltage: 12\,V DC
        \item Flow Rate: $>15.0$\,L/min
        \item Vacuum Pressure: $-60$\,kPa
        \item Power: 12\,W
    \end{itemize}
    \item \textbf{MCU:} Arduino Uno R3
    \item \textbf{Solenoid Valve:} Generic 0520F 12\,V DC
    \item \textbf{Suction Setting:} Silicone suction cup (15 mm diameter), metal connector (60 mm length)
    \item \textbf{Mounting Parts:} 3D printed
    \item \textbf{Others:} USB Relay, Silicone tube, 12V lithium battery, USB foot switch (for data collection), L298N Motor Driver
\end{itemize}

\textbf{System control.} We employ the USB protocol to control relays for generating distinct signal codes. Upon receiving a "turn-on" command from the computer via the relay, the MCU activates the L298N driver chip through GPIO interfaces, closing the solenoid valve. This action isolates the silicone tube from the atmosphere, establishing an airtight state. Concurrently, the MCU triggers another relay via a GPIO interface to switch on the vacuum pump, thereby initiating the suction operation. Conversely, upon receiving a "turn-off" command, the MCU uses GPIO interfaces to open the solenoid valve, connecting the silicone tube to the atmosphere to release pressure, and simultaneously deactivates the relay to turn off the vacuum pump. The final status of all system devices is subsequently transmitted back to the computer via the UART protocol.

\subsubsection{Prime Actions}
\label{sec:prime_actions}

We discover that many household tasks can be decomposed into three prime actions: \textbf{Suction}, \textbf{Grasp} and \textbf{Move}.

\textbf{Suction} can be appled to various challenging tasks, especially those including large or handleless objects. Unlike other methods \cite{zeng2018robotic, son2025corner}, which only use one suction cup, each gripper is equipped with two suction cups driven by the gripper jaw. Thus, we can adjust the distance between the two suction cups to fit different objects. For example, for large glass slides, we can set the gripper to its maximum stroke and then proceed with suction (wide-area suction). For wallets of square shapes, we can set the gripper to its minimum stroke and then proceed with suction (point suction).

\textbf{Grasp.} For common objects with handles (including cucumbers and bananas), we use the normal two-finger gripping function to grasp them.

\textbf{Move} refers to a series of actions that move a certain part of an object, including \textit{push}, \textit{pull}, \textit{lift} and \textit{press}. For example, to close a drawer and lift the lid of a delivery box, the \textit{push} function and the \textit{lift} function of the gripper are utilized.

Our hybrid end-effector can successfully execute the aforementioned prime actions, and therefore accomplish household tasks by combining these actions.

\subsection{Vision Language Action Model}

\textbf{VLA Formulation.} For learning VLA (Vision-Language-Action) models, we employ a common dual 6-axis-arm manipulation hardware platform. Unlike other mobile manipulation setups, we use a fixed base. The viewpoint includes a fixed top-view camera and two wrist-mounted cameras, one on each arm. The robot's observation at timestep $t$ consists of base and hand visual inputs $V^b_t$, $V^\text{left}_t$, and $V^\text{right}_t$, the state of each robotic arm $s_t \in \mathbf{R}^7$ (including gripper state), and the suction status $f \in \{\text{True}, \text{False}\}$.

Given a language instruction $L$, the objective is to learn an end-to-end policy $\pi(A_t \mid V_t, L)$ that outputs a low-level, executable action chunk $A_t = \{a_t, a_{t+1}, \dots, a_{t+H}\}$, where $a_t \in \mathbf{R}^{16}$ is formed by concatenating the 6-degree-of-freedom joint vectors of the two robotic arms, the gripper opening width, and the suction status.

\begin{figure}[t]
    \centering
    \includegraphics[width=0.47\textwidth]{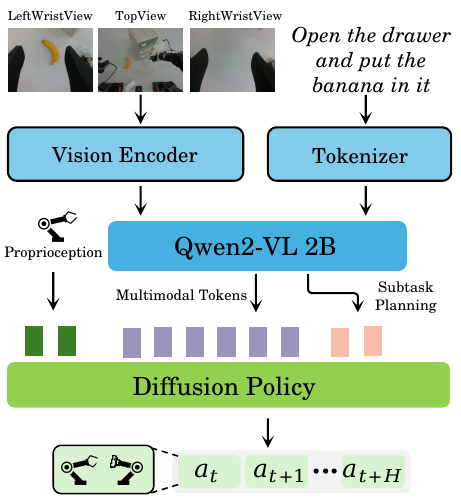}
    \caption{VacuumVLA (based on DexVLA) architecture.}
    \label{fig:fig3}
\end{figure}

\textbf{Shortcut learning in binary suction input.} In existing VLA methods, the current robot state is often included as an observation input. This allows for continuous prediction based on the current state, ensuring that the predicted values within a future action chunk do not deviate too far from the current state. However, the suction status is not well-suited to this paradigm. Our suction status $f \in \{\text{True}, \text{False}\}$ is typically consistent between ground truth and observations across the majority of task learning steps. The status only changes during the specific action chunk when the suction is turned on or off. Such transitions constitute a relatively small proportion of all action chunks in the dataset, making the model prone to the "shortcut" problem (i.e., simply copying the input state). Therefore, in our VLA design, the input is similar to previous VLA approaches, but the output is extended by two dimensions, corresponding to the suction status of the left and right arms, respectively.

\textbf{VacuumVLA} is an end-to-end multimodal robotic policy specifically designed for a suction-gripper hybrid end-effector. To evaluate its effectiveness, we conduct experiments using two distinct state-of-the-art frameworks: $\pi_0$~\cite{black2024pi_0} and DexVLA~\cite{wen2025dexvla}.

When built upon the $\pi_0$ framework, VacuumVLA integrates visual inputs, natural language instructions, and the robot’s proprioceptive state to generate a continuous distribution over actions---including suction and grasping---using a conditional flow matching model. 
In this setup, $\pi_0$ is initialized from PaliGemma \cite{beyer2024paligemma}. 
For action generation, Flow Matching can produce highly precise and realistic outputs. It effectively captures the complex structure and fine details of data, generating motion sequences that are smooth, coherent, and physically plausible.

Alternatively, based on the DexVLA framework---as illustrated in Fig.~\ref{fig:fig3}---VacuumVLA adopts Qwen2-VL as the base vision-language model (VLM). The pretrained image encoder from Qwen2-VL \cite{wang2024qwen2} is applied to project the robot's visual observations---comprising three concatenated images in our setup---into the shared embedding space with language tokens. Unlike $\pi_0$, the VLM generates additional reasoning tokens for subtask planning. Consequently, the diffusion-based action expert can generate action chunks conditioned on three components: multimodal hidden states generated by the VLM, reasoning tokens for subtask planning, and current proprioceptive states. This design enables coherent and hierarchical control.

\section{Experiments}

This section consists of three main parts. The first part is hardware testing, including testing the stability of picking and placing a 500g object and analyzing suction force on different materials. The second part presents a success rate comparison between two versions of VacuumVLA, evaluated on the four predefined gripper primitive tasks. The third part provides visualizations of successful examples.

\subsection{Hardware Function Test}

After designing the hardware structure, we conduct two experiments to test the suction power of our hardware.

\begin{figure}[t]
    \centering
    \includegraphics[width=0.48\textwidth]{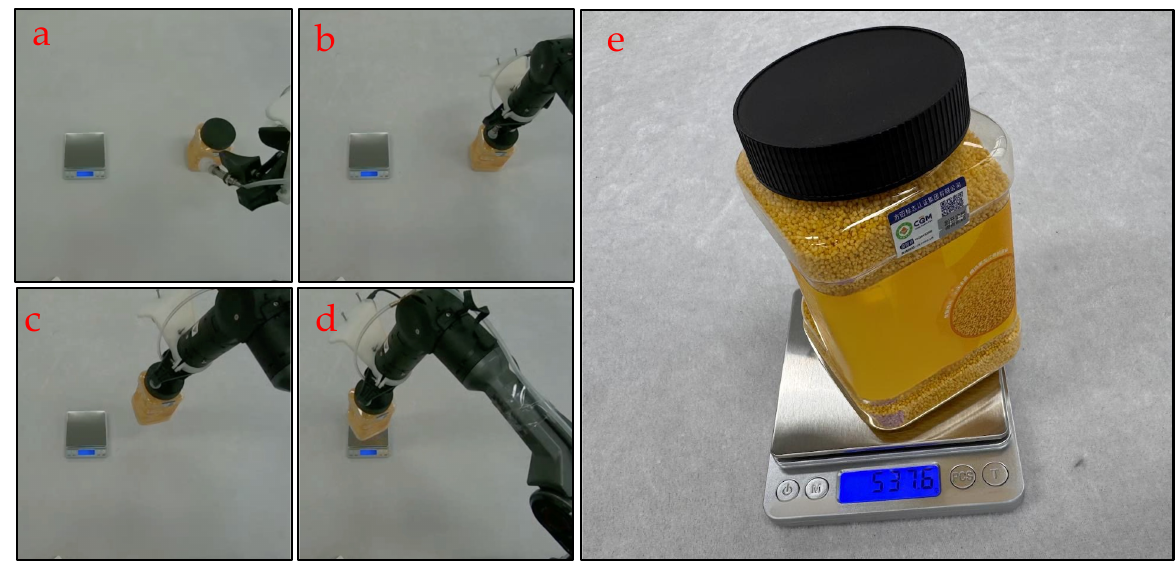}
    \caption{Weight suction test, where figures (a)–(e) correspond to five moments of the test, respectively, and the last image shows an enlarged view of the item placed on the electronic scale.}
    \label{fig:fig4}
\end{figure}

\textbf{Weight suction test.} The first is a 500g object lifting test in Fig. \ref{fig:fig4}. We select a 537-gram jar of rice and test whether it can be successfully picked up from a desktop and placed onto an electronic scale.

\begin{figure}[t]
    \centering
    \includegraphics[width=0.48\textwidth]{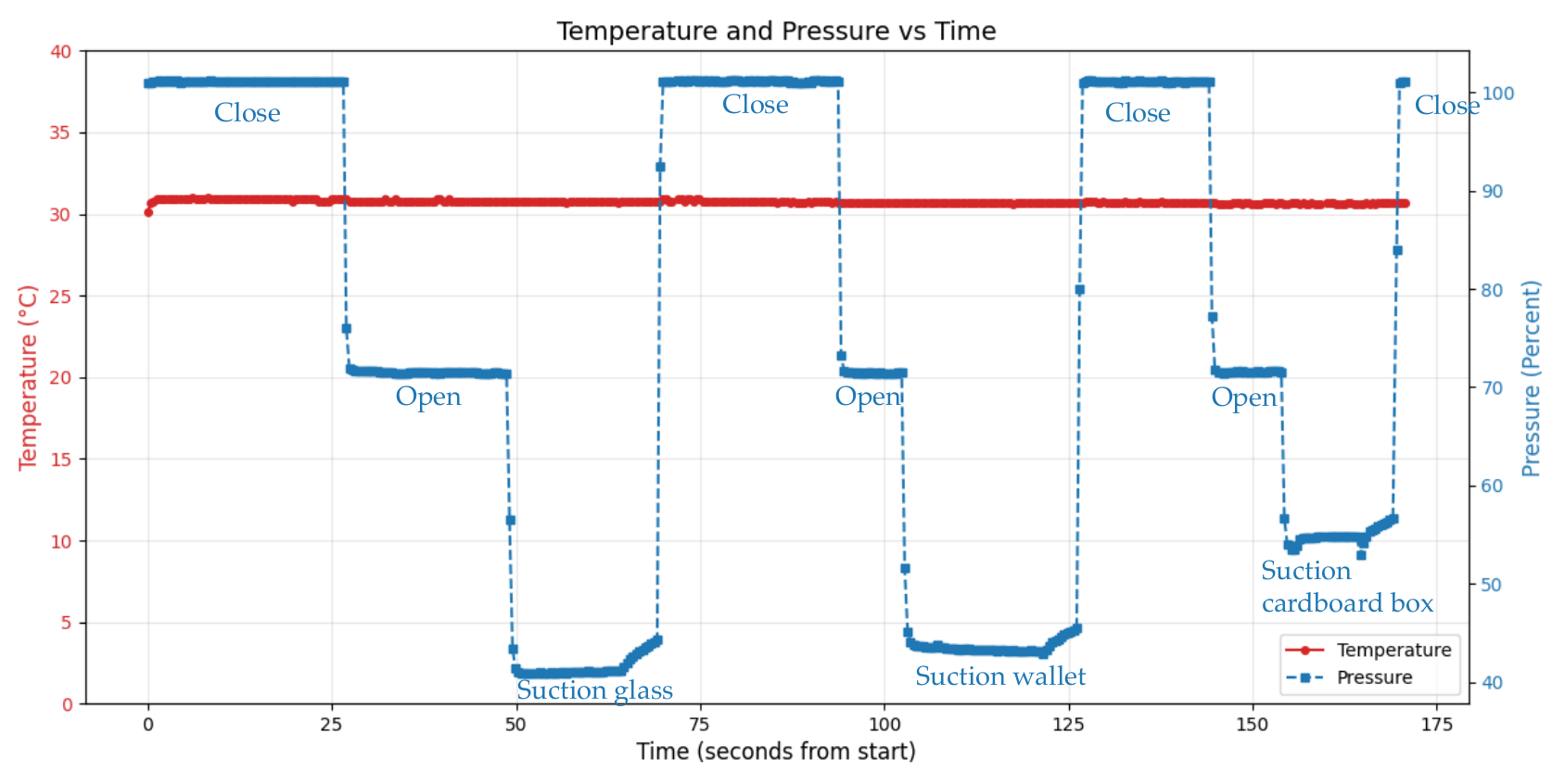}
    \caption{Suction pressure tests for different materials (relative to ambient atmospheric pressure). The ``Close'' phase represents that the motor is turned off. The ``Open'' phase indicates that the motor is on but no object is being sucked. The ``Suction'' phase refers to the system's response when the motor is on and actively sucking different target objects.}
    \label{fig:fig5}
\end{figure}

\textbf{Pressure for different materials.} The second experiment involves pressure tests on different materials. We select glass, a leather wallet, and a cardboard box, with atmospheric pressure at 100 Percent. Through these tests as shown in Fig. \ref{fig:fig5}, we observe that, under the same power setting, the suction capability on porous cardboard material is weaker, which aligns with the general phenomenon that suction cups perform poorly on porous surfaces. (Note: We assume that minor temperature fluctuations over a short period do not significantly affect the air pressure inside the tube.)

\begin{figure}[t]
    \centering
    \includegraphics[width=0.4\textwidth]{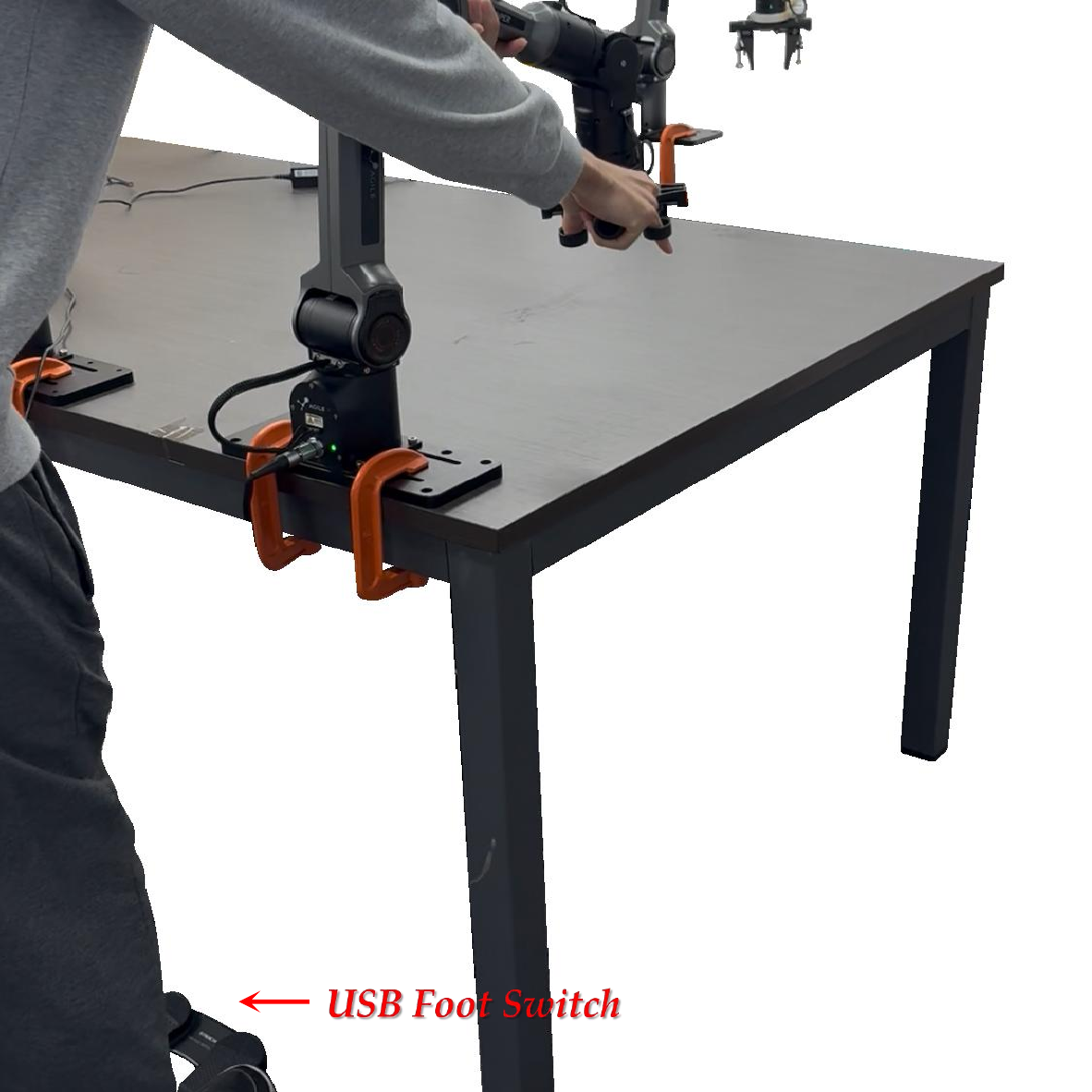}
    \caption{Data collection based on homogeneous teleoperation hardware, where the suction cup is controlled by a foot-operated USB device.}
    \label{fig:trigger_mechanism}
\end{figure}

\begin{figure*}[htbp]
  \centering
  \includegraphics[width=1.0\textwidth]{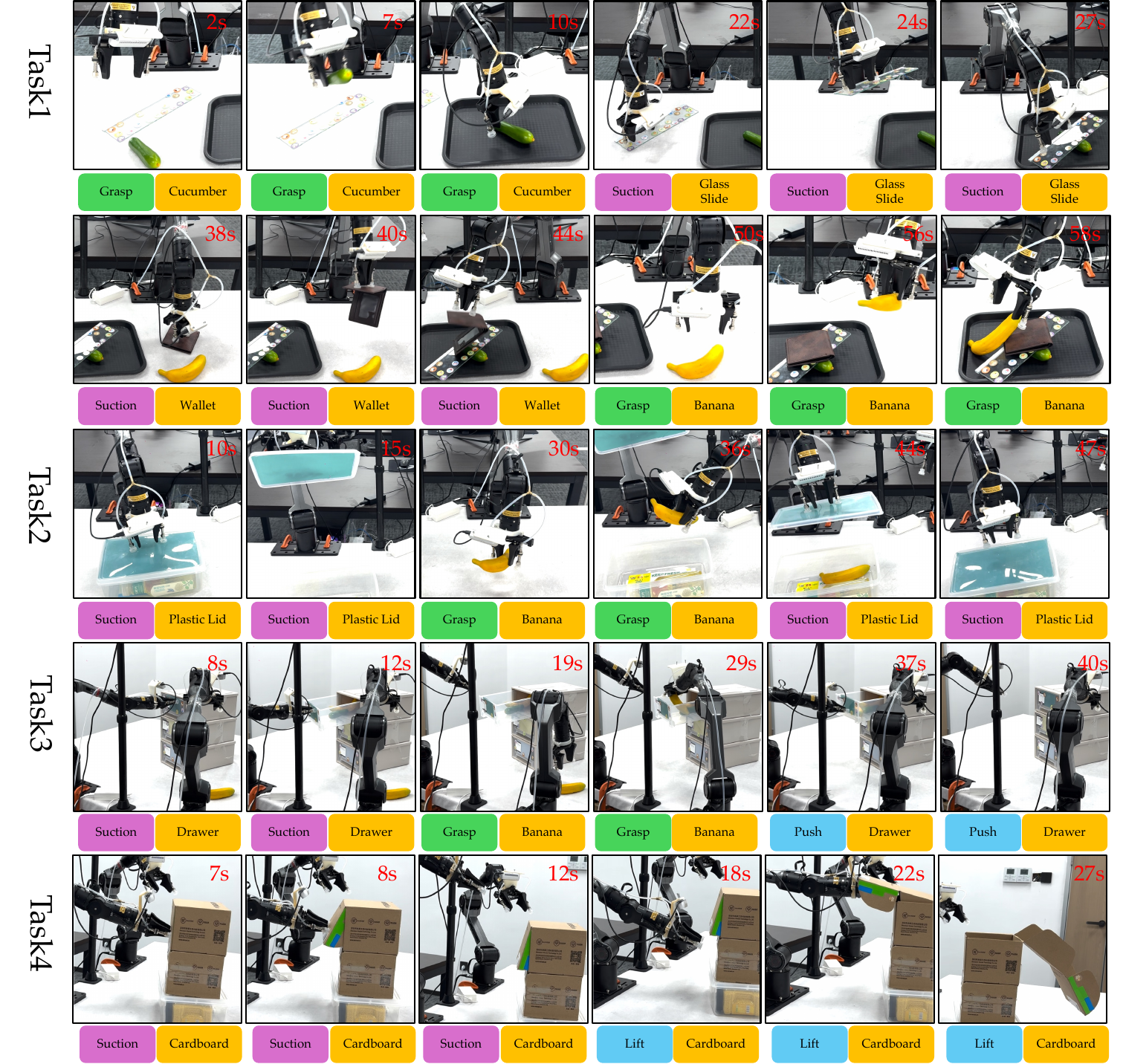}
  \caption{A visualization of the temporal execution of prime-actions and object types across four tasks in VacuumVLA, where prime-action-grasp is shown in green, prime-action-suction in purple, prime-action-lift or push in blue, and object type in yellow.}
  \label{fig:demos}
\end{figure*}

\subsection{VacuumVLA}

As described in the Methods section, we base our experiments on two state-of-the-art VLA frameworks: DexVLA and $\pi_0$. Both DexVLA and $\pi_0$ use training datasets identical in size and diversity, collected through homogeneous teleoperation. The key difference is that DexVLA includes an additional annotation of subtask planning in the form of language instructions, which we elaborate on later.

\textbf{Testing tasks.} We define four long-horizon tasks to test our gripper:
\begin{itemize}
  \item \textbf{Task1}: place the following objects into a tray: a 280\,mm $\times$ 80\,mm glass slide, a banana prop, a cucumber prop, and a wallet.
  \item \textbf{Task2}: open a sealed plastic container, place either the banana or the wallet inside, and close the container.
  \item \textbf{Task3}: open a handleless drawer, place the cucumber inside, and close the drawer.
  \item \textbf{Task4}: open a delivery cardboard box.
\end{itemize}
All actions can be composed of the prime actions described in Section~\ref{sec:prime_actions}, as shown in Fig.~\ref{fig:prime_actions}.

\textbf{Data collection.} For data collection with the robotic arm, we use homogeneous teleoperation. For suction cup data collection, we employ a foot-operated USB switch. When a trigger signal is detected, the system sends a command to turn on the vacuum pump; pressing the trigger again turns it off, as illustrated in Fig.~\ref{fig:trigger_mechanism}. 

The number of trajectories collected varies across different tasks: 200 for Task 1, and 100 each for Task 2, Task 3, and Task 4. During data collection, we switch between suction cup hands.

Since DexVLA requires subtask annotations, we have designed a set of subtask templates for the four tasks. Examples include:

\begin{itemize}
\item \textit{Please use the right arm to suction the glass and place it into the brown dinner plate.}
\end{itemize}



\textbf{Training details.} For the $\pi_0$ model, the batch size is set to 16. However, since the open-source $\pi_0$ is based on JAX and only single-node training code is released, we trained the model for four days until 80,000 steps.

For the DexVLA model, we use the pre-trained one-stage Action Expert. To accelerate computation, we adopt the 400M-parameter variant~\texttt{scale\_dp\_l}\footnote{https://huggingface.co/lesjie/scale\_dp\_l}. Unlike the original DexVLA, which trains in three stages, we train only Stage~2, as our task does not involve cross-embodiment generalization. We do not use LoRA during training. The batch size is 16 per GPU, and training runs on four A100 servers for two days with a constant learning rate of~$2 \times 10^{-5}$.

\textbf{Success rate.} For the four long-horizon tasks, we evaluated the success rates of two versions of VacuumVLA, with each task tested 15 times. A trial is counted as successful only if all prime-actions within the task are successfully completed; otherwise, it is recorded as a failure. Additionally, we observe a typical failure case: if the end-effector of the robotic arm continuously oscillates at the same position for more than one minute, the trial is counted as a failure.

\begin{table}[htbp]
\centering
\caption{Success rates of VacuumVLAs. Traditional end-effectors got success rates of zero primarily due to their requirements for handles and proper object sizes (Section~\ref{sec:limitations_of_previous} for more details).}
\label{tab:VacuumVLA_success}
\begin{tabular}{|c|c|c|c|c|c|c|}
\hline
\textbf{Model Base} & \textbf{End Effector} & \textbf{Task1} & \textbf{Task2} & \textbf{Task3} & \textbf{Task4} \\
\hline
DexVLA & Gripping & 0.0\% & 0.0\% & 0.0\% & 0.0\% \\
\hline
DexVLA & Suction-Gripping & 73.3\% & 80.0\% & 53.3\% & 33.3\% \\
\hline
$\pi_0$ & Suction-Gripping & 53.3\% & 66.67\% & 60.0\% & 53.3\% \\
\hline
\end{tabular}
\end{table}

From the success rate table above, it can be observed that traditional end-effectors fail to handle tasks such as grasping a long glass, opening a handleless plastic container, opening a handleless drawer, and opening
a delivery cardboard box. This failure is primarily due to their requirements for handles and proper object sizes, with detailed analysis presented in Section~\ref{sec:limitations_of_previous}. In contrast, our proposed VacuumVLA achieves reasonable results under two different state-of-the-art VLA frameworks. The detailed visualizations are shown in the following section. 

As shown in Table~\ref{tab:VacuumVLA_success}, we present two variants of VacuumVLA based on different model bases. 
On Task1 and Task2, VacuumVLA with DexVLA as the base model achieves higher success rates than the version using $\pi_0$ as the base. 
However, the opposite trend is observed on Task3 and Task4. 
Specifically, for Task2 (putting the lid on the plastic box), the error offset is 5.3~cm for VacuumVLA (DexVLA) and 3.1~cm for VacuumVLA ($\pi_0$). 
Although the $\pi_0$-based model performs worse in terms of success rate, it achieves higher precision in completing the capping task.

\subsection{Visualizations}

We separately visualize task1, task2, task3 and task4 of VacuumVLA (dexvla) in Fig.~\ref{fig:demos}. 
\footnote{The complete demo can be found at }. 

\section{Limitations}
\label{sec:limitations}
Although the hardware and end-to-end VLA algorithm we designed achieve certain effectiveness in tasks that conventional parallel grippers cannot complete, several issues remain:

\begin{enumerate}
    \item The designed position of the suction cup may interfere with normal grasping in cluttered scenes, as the presence of the suction cup requires a larger distance between target objects.
    \item The two versions of the VLA algorithm proposed in this paper do not account for whether a suction event is a true suction (successful attachment) or a false suction (correct positioning but misaligned suction cup). Since the visual features of true and false suction events are nearly identical, this can lead to the robotic arm performing ineffective motions.
    \item The two suction cups on a single arm proposed in this paper adopt an underactuated design — that is, to simplify the suction end-effector, we use a single motor and a single solenoid valve to control both suction cups on one arm, which may result in one cup achieving proper suction while the other leaks air.
\end{enumerate}

\section{Conclusion}

This paper presents a multifunctional end-effector based on a parallel gripper, integrating both suction and grasping capabilities, enabling the completion of tasks that were previously impossible for a single parallel gripper, such as opening handle-less drawers and cardboard boxes. Furthermore, to validate the proposed multifunctional end-effector, we implement a hybrid suction-grasping VLA (VacuumVLA) based on two existing state-of-the-art VLA frameworks, demonstrating that end-to-end control of combined suction and grasping actions can be achieved through the VLA approach. For future work, we plan to focus on addressing the issues outlined in Section~\ref{sec:limitations}. We anticipate that this work will offer a new perspective on enhancing VLAs' performance through innovative hardware designs.

\addtolength{\textheight}{-12cm}   










\end{document}